\documentclass[10pt,twocolumn,letterpaper]{article}

\usepackage{cvpr}
\usepackage{times}
\usepackage{epsfig}
\usepackage{graphicx}
\usepackage{subcaption}
\usepackage{amsmath}
\usepackage{amssymb}
\graphicspath{ {./images/} }

\usepackage[pagebackref=true,breaklinks=true,letterpaper=true,colorlinks,bookmarks=false]{hyperref}

\cvprfinalcopy % *** Uncomment this line for the final submission

\def\cvprPaperID{406} % *** Enter the CVPR Paper ID here

\ifcvprfinal\fi
\begin{document}

%%%%%%%%% TITLE
\title{Fully Convolutional Neural Networks for Crowd Segmentation}

\author{\\
Kai Kang\ \ \ \ Xiaogang Wang\\
The Chinese University of Hong Kong\\
{\tt\small kkang,xgwang@ee.cuhk.edu.hk}
% \and
}

\maketitle
%\thispagestyle{empty}

%%%%%%%%% ABSTRACT
\begin{abstract}
In this paper, we propose a  fast fully convolutional neural network (FCNN) for crowd segmentation. By replacing the fully connected layers in CNN with $1 \times 1$ convolution kernels, FCNN takes whole images as inputs and directly outputs segmentation maps by one pass of forward propagation. It has the property of translation invariance like patch-by-patch scanning but with much lower computation cost. Once FCNN is learned, it can process input images of any sizes without warping them to a standard size. These attractive properties make it extendable to other general image segmentation problems. 

Based on FCNN, a multi-stage deep learning is proposed to integrate appearance and motion cues for crowd segmentation. Both appearance filters and motion filers are pre-trained stage-by-stage and then jointly optimized. Different combination methods are investigated. The effectiveness of our approach and component-wise analysis are evaluated on two crowd segmentation datasets created by us, which include image frames from $235$ and $11$ scenes, respectively. They are currently the largest crowd segmentation datasets and will be released to the public. 
\end{abstract}

\section{Introduction} % (fold)
\label{sec:intro}
Crowd video surveillance in public areas with high population density has drawn a lot of attentions because its important applications in public security and traffic management. Related research topics include crowd segmentation \cite{TaoZhao:2003bayesian,Dong:2007shapeindexing,Shah:2007hi,Chan:2008dtm,Zhao:2008segmentation,Arandjelovic:2008vf}, people counting \cite{Kong:2005wa,Chan:2012counting,Loy:2013dt,Seibert:2013bt,Ma:2013crossing}, crowd tracking \cite{Zhao:2008segmentation,Li:2009learning}, crowd behavior analysis \cite{Solmaz:2012ux,Zhou:2012hx,Zhou:2013ko,Shao:2014scene}, and abnormality detection \cite{Yi:2014L0}. Among them, crowd segmentation is a fundamental problem serving as the basis of other crowd related techniques. It has significant influence on the performance of other tasks. For example, some people counting methods \cite{Kong:2005wa,Chan:2012counting,Ma:2013crossing} require crowd regions being segmented first. Before tracking crowds or analyzing their behaviors, their locations need to be known in advance. 

\begin{figure}[t]
    \centering
    \includegraphics[width=\linewidth]{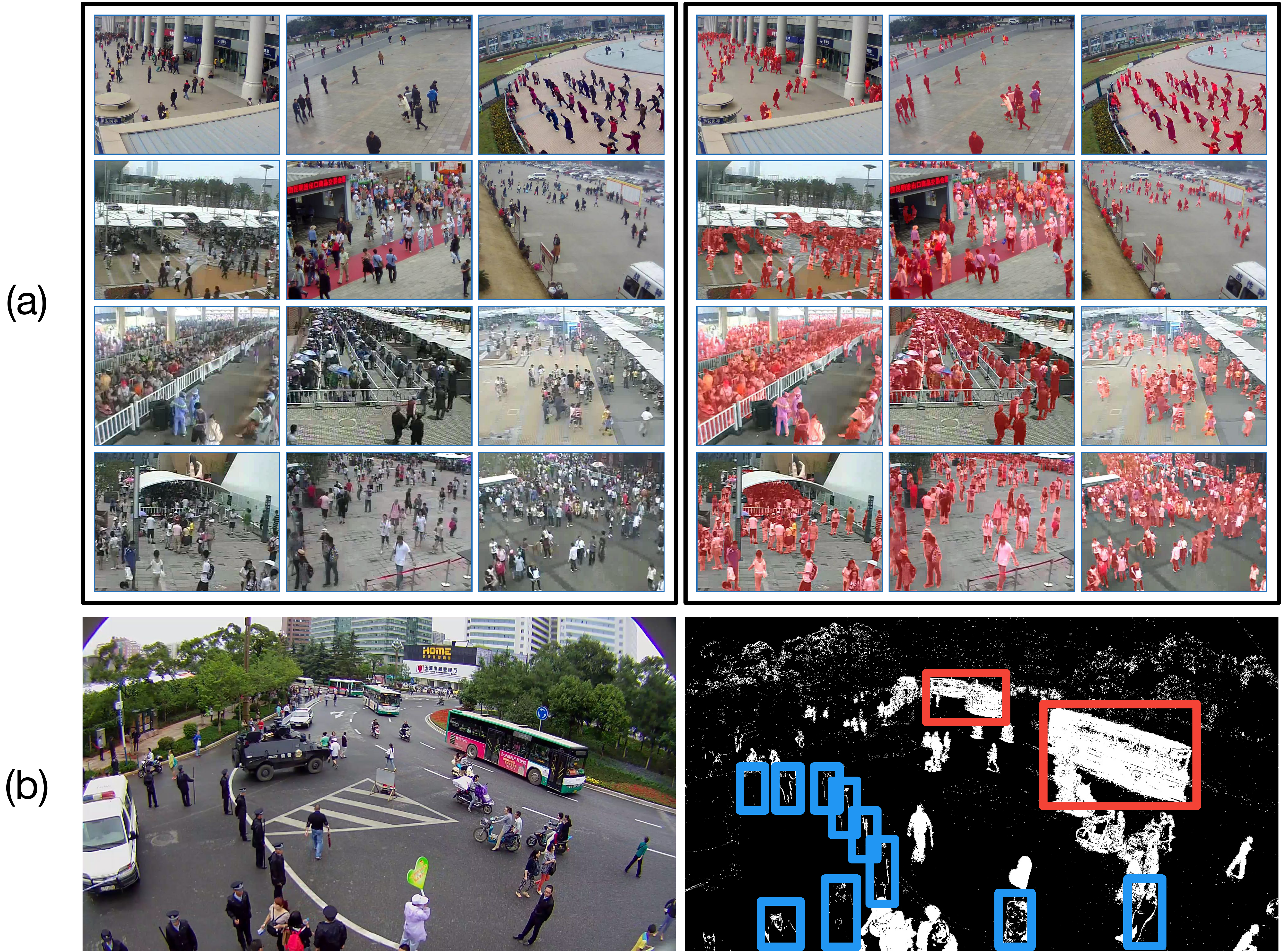}
    \caption{(a) Real world crowd segmentation. (Left) the real-world scenes with large differences in perspective distortion, crowd and background appearance, weather and lighting conditions. (Right) the ideal segmentation results (shown as red overlay). (b) Motion segmentation. (Right) Motion segmentation results using Gaussian mixture model. The blue boxes are stationary groups failed to be detected (false negatives), and red boxes indicate moving objects that are falsely detected (false positives). Best viewed in color.}
    \label{fig:motivation}
\end{figure}

However, crowd segmentation is a challenging problem. A commonly used solution in practice is through motion segmentation assuming that the surveillance cameras are static. However, as examples shown in Figure~\ref{fig:motivation} (b), in many scenarios there are large groups of people being stationary for a long time and they cannot be captured by motion cues. Yi et al. \cite{Yi:2014profiling,Yi:2014L0} showed that such stationary crowds are worthy of special attention since they have large influence on the traffic flows. Moreover, some moving objects of other categories might be detected by motion as false positives. 

It is important to extract appearance cues for crowd segmentation. Due to heavy occlusions among people and small pedestrian sizes in crowd, pedestrian detectors generally do not work well for crowd. As shown in Figure~\ref{fig:motivation} (a), the appearance of crowds and their background change significantly across different scenes and are under different perspective distortions. Ideally, the training data should not have overlap with the target scenes in the test set, i.e., users are not expected to label training samples from the target scenes. Appearance-based crowd segmentation should have invariance across scenes. Widely-used handcrafted features such as HOG \cite{Dalal:2005hog}, SIFT \cite{Lowe:2004kp} and LBP \cite{ojala2002multiresolution} cannot handle such complex variations well.

Deep learning has achieved great success in many computer vision problems in recent years. However, to the best of our knowledge, no feature representations have been specifically learned for human crowds with deep models yet. Integrating appearance and motion information into the deep learning model could increase the accuracy of crowd segmentation.

\subsection{CNN for segmentation}
\label{ssec:ccnseg}
\begin{figure*}[t]
    \centering
    \includegraphics[width=\textwidth]{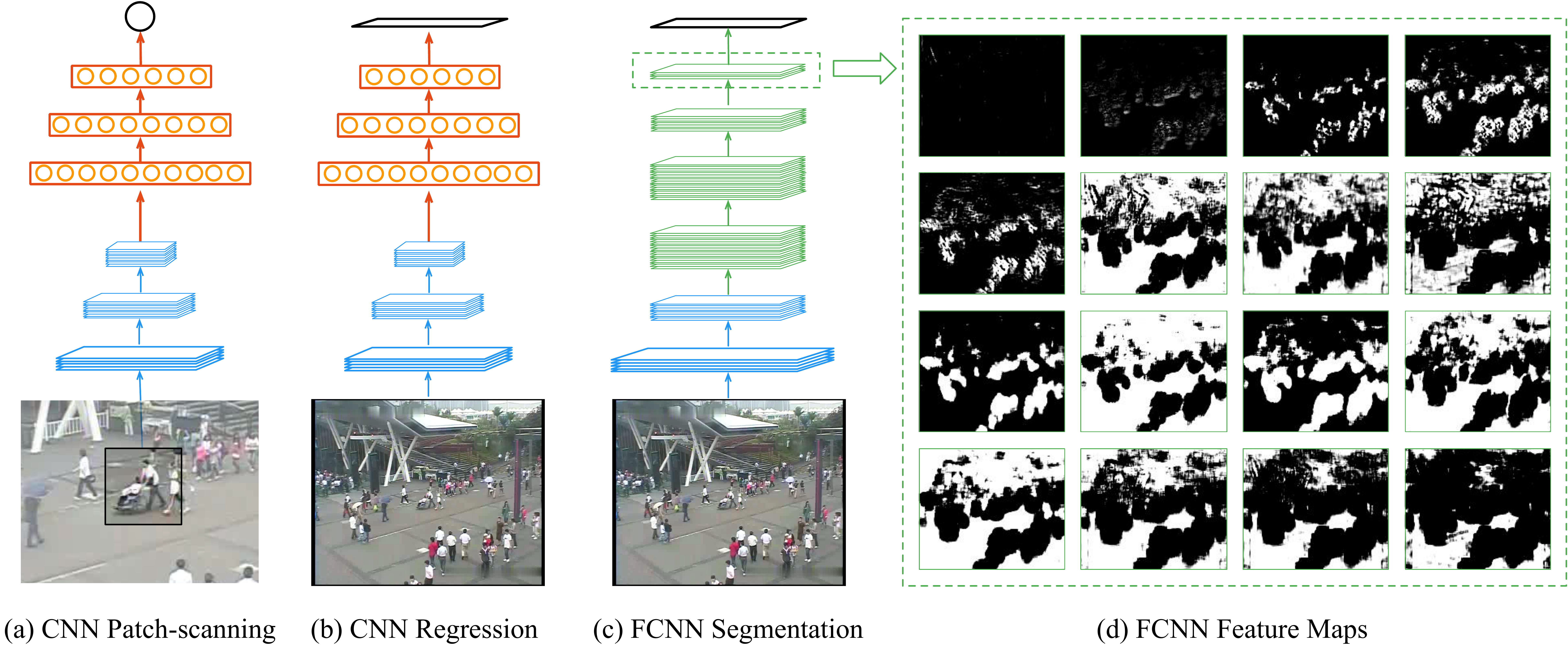}
    \caption{CNN and FCNN models. Blue, red, and green layers represent convolutional-pooling layers, fully-connected layers, and ``fusion'' convolutional layers in FCNN, respectively. (a) A typical structure for patch-by-patch scanning method. (b) A typical CNN structure for regression problem. (c) Our proposed FCNN network. (d) The output maps of a high-level fusion layer in FCNN. The upper $2$ rows capture some edge and texture information, while the lower $2$ rows capture more context information.}
    \label{fig:model_comprison}
\end{figure*}

The Convolutional neural network (CNN) is widely used in computer vision. Its typical structure is shown in Figure~\ref{fig:model_comprison} (a). Following multiple convolutional and pooling layers, which extract features, several fully-connected layers predict the class labels. As shown in Figure~\ref{fig:model_comprison} (a), the typical way of applying CNN to image segmentation is via patch-by-patch scanning \cite{Farabet:2013Learning,Pinheiro:2014recurrent}. To predict the class label of a pixel, its surrounding patch is cropped and fed to the CNN. Studies \cite{Farabet:2013Learning} showed that large patch sizes generally lead to better segmentation accuracy, since large patches capture more contextual information, which can be well learned by deep models. This patch-by-patch scanning approach has translation invariance, i.e., the prediction of a pixel label only depends on its surrounding patch and is independent of its coordinates. Therefore, it has been widely adopted in CNN based image segmentation. The major drawback is its computational cost, since the forward propagation has to be repeated for $N$ times, where $N$ is the number of pixel labels to be predicted. Our experiment shows that it takes five minutes to obtain pixel-level segmentation on a frame of size $576 \times 720$ with GPU implementation, which is impractical for real-time surveillance applications. 

An alternative way is to input the whole images to CNN, which directly predicts the whole segmentation maps with the last fully connected layer (Figure~\ref{fig:model_comprison} (b)). With only one pass of forward propagation, its computation cost is low. The last fully connected layer, however, essentially learns different classifiers for different locations. Therefore, it does not have translation invariance. It is also required that the input images must be fixed in size, because the fully connected layers require their inputs to be fixed sizes. This approach was only applied to normalized images with regular structures such as pedestrians \cite{Luo:2013pedestrian}, but is not suitable for general image segmentation.  Both limitations come from the existence of fully connected layers. 
If a deep model only has convolution layers and pooling layers, it should have translation invariance, since both convolutional filters and pooling kernels are translation invariant, i.e., their outputs only depend on surrounding regions but not locations. Once filters at multiple layers have been learned, they can be applied to images of any sizes. The output feature maps of such CNN models vary in sizes, which are proportional to the sizes of input images. 

\subsection{Our method}
In this work, we propose a novel fully convolutional neural network (FCNN) to learn both appearance features and motion features for crowd segmentation. As shown in Figure 2 (c), FCNN removes all the fully connected layers in CNNs and places $1 \times 1$ convolution kernel in the last year to predict labels at all the pixels in the output segmentation map. It integrates the advantages of both CNN based segmentation methods described in Section \ref{ssec:ccnseg}. 1) FCNN takes whole images as inputs and directly outputs the whole segmentation maps with one pass of forward propagation.  It takes 125ms to segment one frame of size $576 \times 720$, $2400$ times faster than path-by-patch scanning. 
2) FCNN has translation invariance, since it is only composed of convolutional layers and pooling layers. The prediction at a pixel in the segmentation map only depends on its surrounding region in the input  image. With six convolutional layers and two pooling layers, its receptive field is quite large in the input image and therefore much contextual information can be captured, which leads to good segmentation accuracy. 3) The input images can be of arbitrary sizes. Training and test images could be of different sizes. Therefore, image size normalization  is not a required preprocessing, which avoids distortion to the images.

We propose multi-stage deep learning to combine appearance, motion and structure cues for crowd segmentation. It is not a good choice to directly combine them as different input channels or average their decisions as two separate classifiers, because they have difference roles. For example, if appearance cues provide enough confidence on a patch being crowd, we will label it as crowd even though it has no motion, because some crowds could be stationary. On the other hand, even if a patch has strong motion, it still needs to be verified by  appearance cues, because there are many other types of moving objects in the world. In the proposed multi-stage deep learning, the motion filters are pre-trained only with samples which cannot be confidently classified by appearance features. The class labels are predicted by taking the output of appearance filters as contextual information. Appearance filters and motion filters are pre-trained state-by-stage and then jointly optimized by fine-tuning.

The effectiveness of the proposed model is evaluated on two large scale crowd segmentation datasets created by us. They include 235 scenes from Shanghai World Expo 2010 and 11 scenes from a city respectively. A total of $7994$ frames are manually annotated with ground truth. Detailed component-wise analysis is provided. 
% It is also compared with state-of-the-art on the UCSD public dataset. 

% section intro (end)

\section{Related Work} % (fold)
\label{sec:related_work}
A number of methods have been proposed for crowd segmentation in recent years. It is typically achieved via background subtraction and motion segmentation \cite{Zivkovic:2004improved,Chan:2008dtm,Dong:2007shapeindexing,Mumtaz:2014joint}, which usually require a static camera view or fixed pedestrian motion patterns. 
% 
% Some also use pedestrian detection and tracking methods \cite{Viola:2005detecting,Dalal:2005hog,Leibe:2005pedestrian,Zhao:2008segmentation},
Some approaches based on pedestrian detection and tracking results \cite{Viola:2005detecting,Dalal:2005hog,Leibe:2005pedestrian,Zhao:2008segmentation} usually perform poorly on highly crowded scenes due to severe occlusions. Combining multiple visual cues into crowd segmentation has also been investigated by using motion and shape information jointly \cite{Rittscher:2005simultaneous,Dong:2007shapeindexing}. Most of these works require training and testing on the same scene, which is not applicable for real-world cross-scene segmentation.

Deep neural networks have been widely deployed in general image segmentation or scene labeling tasks. 
% A typical method for image segmentation is through patch-by-patch scanning\cite{Farabet:2013Learning,Pinheiro:2014recurrent,Ciresan:2012deep}, or fully-connected layer regression\cite{Luo:2013pedestrian}. Due to the fully-connected layers, these methods requires a fixed size of input and output.
The traditional methods for image segmentation are patch-by-patch scanning \cite{Farabet:2013Learning,Pinheiro:2014recurrent,Ciresan:2012deep} and fully-connected layer regression \cite{Luo:2013pedestrian}, which requires the fixed size of input and output.

Lin et al. \cite{Lin:2013nin} have employed feature maps instead of fully-connected layers to increase network representation capability. The ``mlpconv'' layers in their proposed network, however, behave similar to patch-by-patch scanning. $1 \times 1$ convolutional kernels have recently been extensively used in GoogLeNet \cite{Szegedy:2014goingdeeper} to reduce computation cost and thus increase the network sizes in depth and width.

% section related_work (end)

\section{Fully Convolutional Neural Networks} % (fold)
\label{sec:fcnn}
\subsection{Convolutional Neural Networks (CNN)} % (fold)
\label{sub:cnn}

Typical convolutional neural networks \cite{LeCun:1989backpropagation,LeCun:1998gradient} usually consist of convolutional layers, pooling layers, neuron layers, and fully-connected layers. 

\vspace{0.1cm}\noindent 1) \textit{Convolutional layers} convolve the input image or feature maps with a linear filter and can be denoted as
\begin{equation}
    (h_k)_{ij}=(W_k*x)_{ij}+b_k
    \label{eq:conv}
\end{equation}
where $k=1,...,K$ denotes the index of neuron, $x$ denotes the input feature maps, $W_k$ and $b_k$ are the $k$-th filter and bias, and $(h_k)_{ij}$ is the $(i,j)$ element of the $k$-th output feature map. ``$*$'' denotes the $2$-D spatial convolution. The output feature maps represent the responses of each filter $W_k$ on the input image or feature maps.

\vspace{0.1cm}\noindent 2) \textit{Pooling layers} are non-linear down-sampling layers that yield maximum or average values in each sub-region of input image or feature maps. Pooling layers increase the robustness of translation and reduce the number of network parameters.

\vspace{0.1cm}\noindent 3) \textit{Neuron layers} apply nonlinear activations on input neurons. Common activations are identical function, sigmoid function, hyperbolic tangent function, rectified linear unit, \etc.

\vspace{0.1cm}\noindent 4) \textit{Fully-connected layers} compute outputs by
\begin{equation}
    y_k=\sum_l{W_{kl} x_l}+b_k
    \label{eq:fc}
\end{equation}
where $x_l$ is the $l$-th input neuron, $y_k$ is the $k$-th output neuron, $W_{kl}$ denotes the weight connecting $x_l$ with $y_k$, and $b_k$ is the bias term of $y_k$. The parameters of fully-connected layers are the weight matrix $W$ and bias vector $b$, which require fixed numbers of inputs and outputs as mentioned in Section~\ref{sec:intro}. 

Nevertheless, Equation~\eqref{eq:fc} can be re-written as
\begin{equation}
    (y_k)_{1,1}=(W_k * x)_{1,1}+b_k
    \label{eq:fc_as_conv}
\end{equation}
where ``$*$'' is $2$-D spatial convolution same as that in Equation~\eqref{eq:conv} while $y$ and $x$ are $1 \times 1$ feature maps. In this way, we can treat fully-connected layers as convolutional layers with $1 \times 1$ kernels, which are used in proposed fully-convolutional neural networks.
% subsection cnn (end)

\subsection{Fully-convolutional Neural Networks (FCNN)} % (fold)
\label{sub:fcnn}
In CNNs, convolutional layers and pooling layers perform local operations on the input feature maps. These operations preserve the spatial relationships of neighboring neurons and are unrelated to the shape of input feature maps. These properties are desirable in segmentation tasks for that segmentation essentially performs the same detection or classification operation on each pixel of the input images.

As shown in Equation~\eqref{eq:fc} and Equation~\eqref{eq:fc_as_conv}, the fully-connected layers in traditional CNNs are equivalent to convolutional layers with $1 \times 1$ kernels, which in fact perform a ``fusion'' operation on the input feature maps. The output fusion feature maps keep the same dimensions as input feature maps, 
each of which captures different information, as shown in Figure~\ref{fig:model_comprison} (d).
% while each capturing different information (Figure~\ref{fig:model_comprison} (d)).

Figure~\ref{fig:model_comprison} (c) shows our proposed FCNN architecture for crowd segmentation. The inputs are original frame images and ground-truth segmentation maps. The frame images are fed into the FCNN with two convolutional-pooling layers and several fusion layers. The output feature map represents the probability of being crowd on each down-sampled pixel. With proper padding, all convolutional and fusion layers keep the same dimensions with the input feature maps. Each of two pooling layers with $2 \times 2$ non-overlapping regions reduces the dimension by a factor of $2$. The ground-truth segmentation maps are fed into two average-pooling layers to have the same dimension as the output feature maps of FCNN.

The cross entropy is used as the loss in our objective function, which is suitable for binary classification problems like crowd segmentation. As shown in Equation~\eqref{eq:cross_entropy}, the $N$ denotes the number of output neurons, $t_n$ is the target probability from the pooled segmentation map and $o_n$ is the output probability prediction from the FCNN.
\begin{equation}
    E=\frac{-1}{N}\sum_{n=1}^{N}{t_n\log{o_n}+(1-t_n)\log{(1-o_n)}}
    \label{eq:cross_entropy}
\end{equation}

By using the whole frames as training samples and the segmentation maps as labels, the FCNN is globally trained and produces full-frame segmentation results.

% subsection fcnn (end)

\subsection{Receptive Fields of FCNN} % (fold)
\label{sub:receptive_fields}

In CNNs, the receptive fields represent the sensitive regions that affect the output of a neuron. 
% Without loss of generality, 
For the purpose of generalization, we assume 1) the kernels of convolutional and pooling layers are square, 2) the convolution stride is $1$, and 3) the pooling operations perform on non-overlapping regions. Let $N^c$ denote the convolutional kernel size and $N^p$ denote pooling kernel size. The receptive field of a convolutional-pooling layer has the size of $R$, i.e.,
\begin{equation}
    R=N^c+N^p-1
\end{equation}

Receptive fields of neighboring output neurons have a offset of pooling size $N^p$. Convolutional layers without pooling are equivalent to $N^p=1$. The $1 \times 1$ convolutional layers keep the same receptive field as the previous layer. In crowd segmentation, increasing the receptive fields can incorporate more context information into the final prediction. For this purpose, in the fusion layers, we also use kernels of size $3$.
% subsection receptive_fields (end)

%%%%%%%%%%%%%%%%%%%%%%%%%%%%%%%%%%%%%%%%%%%%%%%%%%%%%%%%%
\section{Multi-stage FCNN} % (fold)
\label{sub:multistage}
\subsection{Visual Cues for Crowd Segmentation} % (fold)
\label{sub:visual_cues}

% To combine more information into crowd segmentation and improve the overall performance, we analyze visual cues.
To improve the segmentation performance, we exploit visual cues from both appearance and motion aspects.

% A straightforward visual cue is \textbf{appearance}, which is essentially the input image itself. 
\vspace{0.1cm} \noindent \textit{Appearance}: In recent years, CNNs have achieved various state-of-the-art results in image classification, segmentation, and object detection tasks \cite{Krizhevsky:2012imagenet,Farabet:2013Learning,Ciresan:2012deep,Sermanet:2013overfeat,Girshick:14rcnn,Razavian:2014cnn}. Even for large-scale video classification tasks, researches have shown 
% effectiveness of single frame
the effectiveness of CNNs with single frames as inputs \cite{Karpathy:2014largescale}. For a single frame with a clear background, it is easy and efficient to detect the crowds. The appearance of pedestrians has a visible difference from that of background objects and structures.

\vspace{0.1cm} \noindent \textit{Motion}: 
% especially in video surveillance. Motion information, on the one hand, helps crowd segmentation by detecting the moving pedestrians that have similar textures as background and eliminating background structures such as trees, fences, buildings \etc. On the other hand, motion information usually fails in detecting stationary crowds, where appearance model alone may have good performance.
Although appearance-based models can perform well in detecting stationary crowds, motion-based model can improve crowd segmentation, especially for moving crowds. This is because the motion cues can help detect moving pedestrians that have similar textures with background, and help eliminate background structures, such as trees, fences, buildings, \etc.

\vspace{0.1cm} \noindent \textit{Structure}: 
In fact, the background structures, scene dimensions, and perspective distortions can also help to detect crowds.
For example, it is difficult to detect crowds in the scene with parallel structures or without structures like buildings and floors.
% In addition to visual cues for crowds themselves, the understanding of the background \textbf{structures}, the dimensions of the scenes and perspective distortions also help to find crowds. 
% For example, if an area with parallel structures or no structures at all may be buildings or pure floors, so the chance of find crowd there is low. 
{We use the edge detection results from \cite{Dollar:2013structured} as inputs to the network.} The edge models provide information about what kind of structures are crowd-like versus background-like.
% subsection visual_cues (end)

\subsection{Fusion Schemes and Extensible Architecture} % (fold)
\label{sub:fusion_schemes}
% With multiple visual cues as input, a proper fusion scheme to combine them is necessary. We investigate three fusion schemes: 1) input fusion, 2) feature fusion and 3) decision fusion.
To combine multiple visual cues, we investigate three fusion schemes: input fusion, feature fusion, and decision fusion.
The input fusion directly concatenate input maps as multiple channels. The feature fusion combines output feature maps of a certain fusion layer and use feature maps of all three networks to make a decision. The decision fusion scheme combines the output maps of three separately trained networks. It is similar to learning from multiple experts.

\begin{figure}[t]
    \centering
    \includegraphics[width=\linewidth]{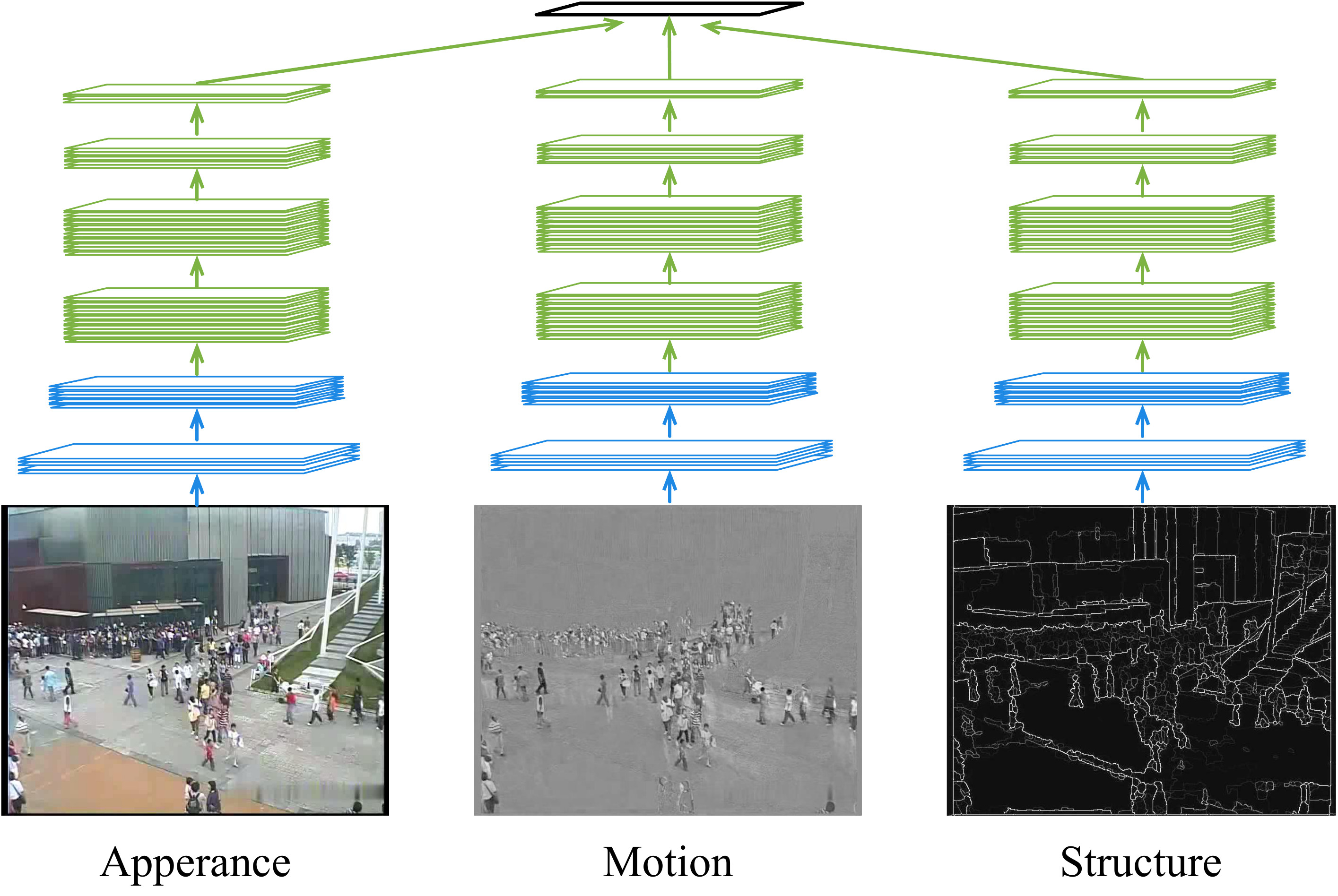}
    \caption{Multi-stage fusion structure. Each branch is pre-trained with different visual cues. The multi-stage combination is done by first fixing appearance branch and fine-tuning motion branch. Then fix appearance and motion branches and fine-tune structure branch. Finally, the whole network is globally fine-tuned.}
    \label{fig:multi_stage}
\end{figure}

For maximum extensibility, we train the two fusion schemes in a cascaded way similar to \cite{Zeng:2013multistage}. For each new branch, we fix the parameter of previous branches and finetune only the new branch (Figure~\ref{fig:multi_stage}). This scheme has two advantages: 1) removing the new branch will not affect the performance of original network and 2) it forces the new branch to learn complementary information to the original network. With this cascaded architecture, we could add more information to improve the performance of the whole system.

\section{Experiments} % (fold)
\label{sec:experiments}
\subsection{Datasets} % (fold)
\label{sub:datasets}
Training and evaluating the proposed FCNN model require a large amount of labeled samples. There are following requirements: 1) the training and test sets should have a large number of distinct camera views, and the two sets should not contain the same camera views; 2) The training set should contain a large amount of labeled frames indicating the localization of pedestrians; 3) The test set should ideally contain pixel-level segmentation ground-truth; 4) To utilize motion information, the dataset should be videos.

As far as we know, however, there is no public datasets available that satisfy those requirements. Therefore, we build two datasets for this work.

\begin{figure}[t]
    \centering
    \includegraphics[width=\linewidth]{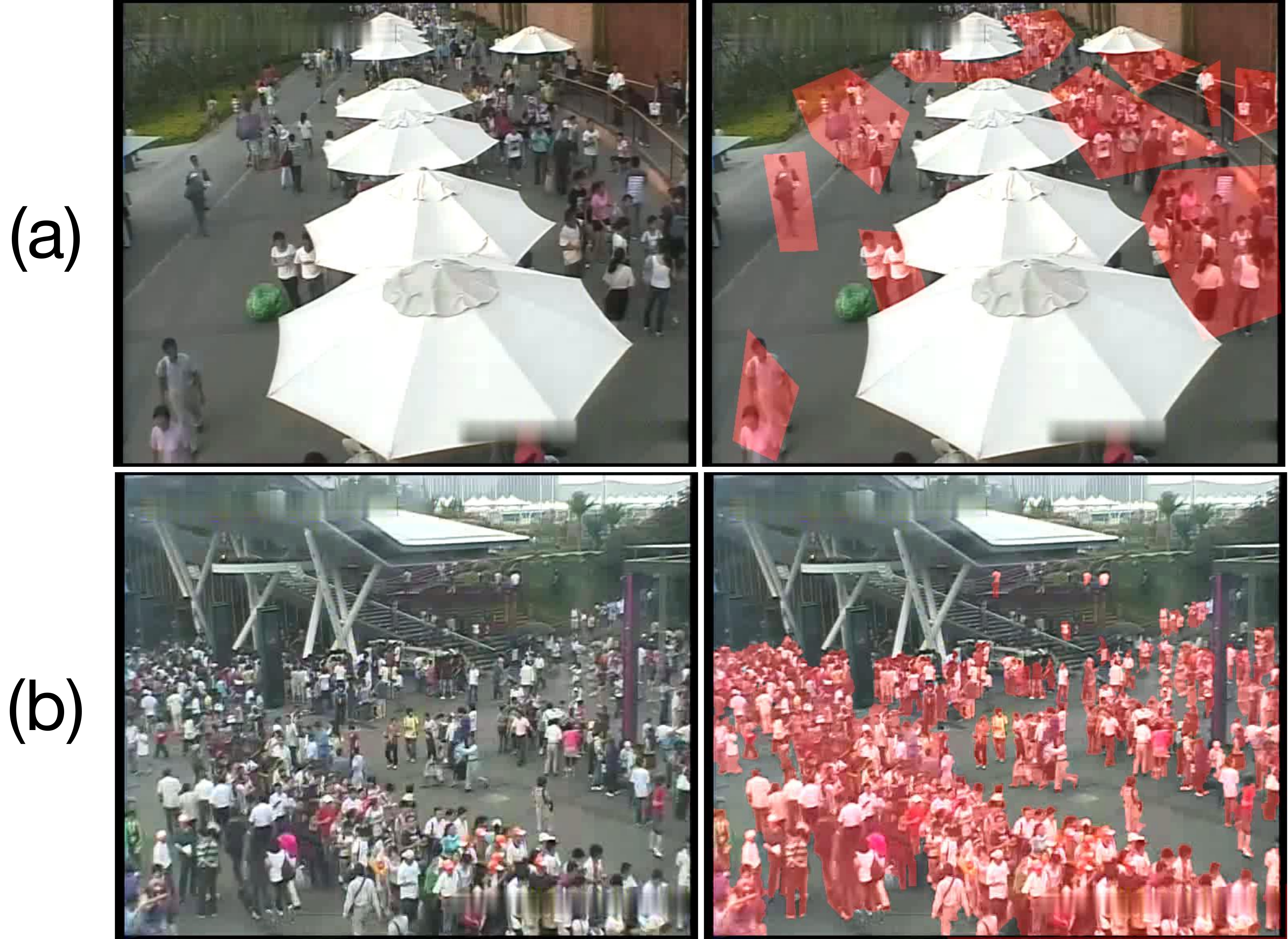}
    \caption{Labeling schemes: (a) the region-level ground truth labels for training and validation sets. The polygons (red) cover the regions of crowds. (b) the pixel-level ground truth for the test set.}
    \label{fig:labeling}
\end{figure}

\vspace{0.1cm}\noindent\textbf{Shanghai World Expo Dataset.} The first dataset contains $235$ camera views collected from $2010$ Shanghai World Expo. $184$ camera views are randomly selected as the training set and the rest $51$ for testing. We extract $6153$ $5$-second video clips from the training set and label $1$ frame in each clip using polygons (Figure~\ref{fig:labeling} (a)). The polygons cover the crowd or pedestrian regions, which provide rough segmentation maps. For the test set, we select $10$ frames for each camera view and label the frames at pixel level (Figure~\ref{fig:labeling} (b)) for accurate full-frame evaluation. 

\vspace{0.1cm}\noindent\textbf{City Dataset.} The second dataset contains $11$ scenes from surveillance cameras in public places, including parks, squares, railway stations, bus stops, streets, \etc. We use this dataset only for testing, in which the cross-scene feasibility is tested in a practical way. The labeling scheme is at pixel-level same as test set in the first dataset.

\begin{figure}[t]
    \centering
    \includegraphics[width=\linewidth]{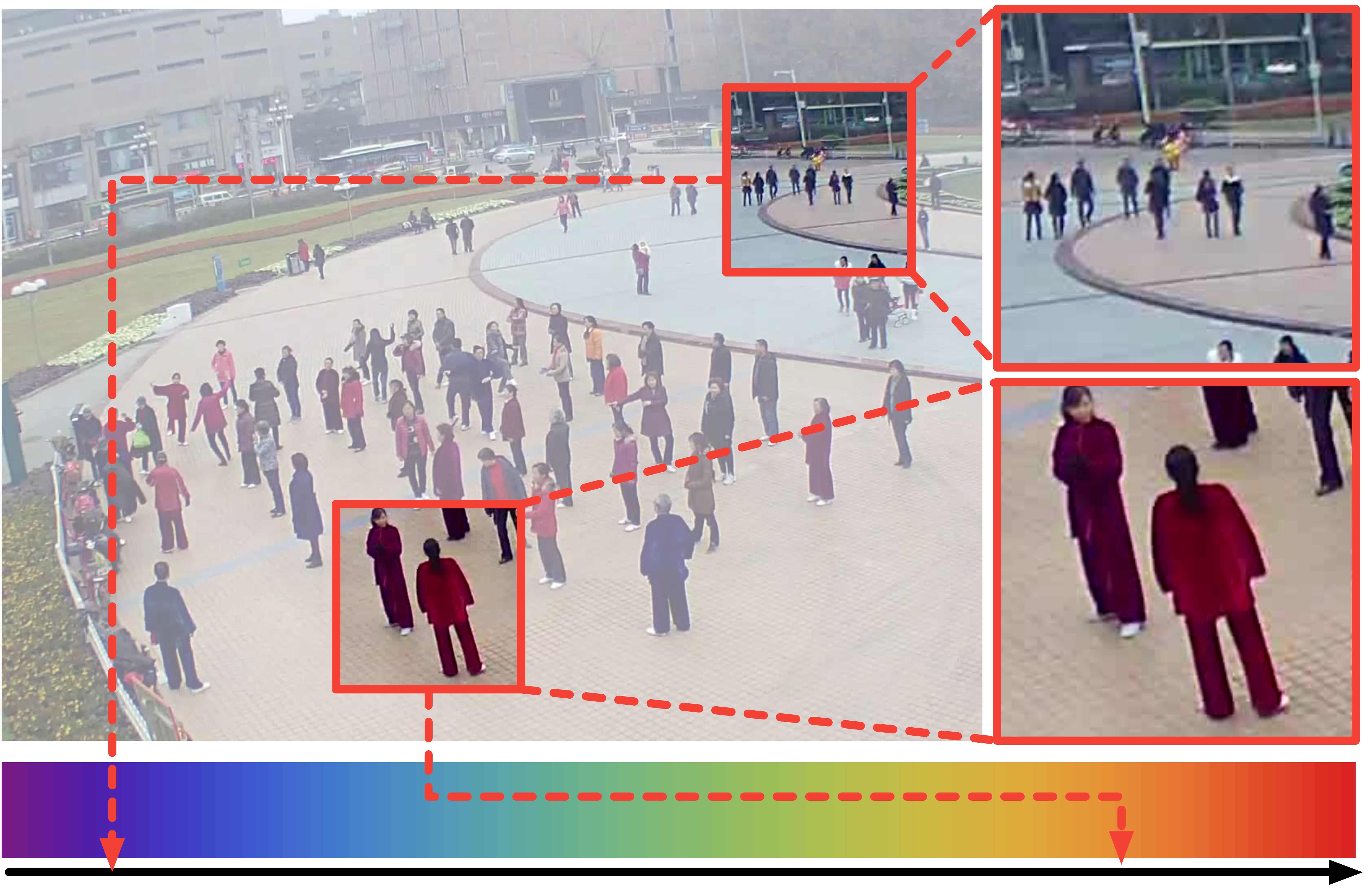}
    \caption{Perspective distortion. Patches at different perspective level (black axis indicates the perspective values) contain significantly different information (two red boxes on the right). The FCNN is trained on $184$ different scenes that cover a wide range of perspective scales.}
    \label{fig:perspective_distortion}
\end{figure}

As shown in Figure~\ref{fig:perspective_distortion}, the appearance of crowd patches is significantly affected by perspective distortion. If a network is trained on insufficient scenes, it could not have enough generalization capability. Our FCNN model has been trained on $184$ different scenes that provide large perspective variety.

\subsection{Baseline Models} % (fold)
\label{sub:baselines}
In comparison with our proposed method, we test several baseline models.

The first baseline model is the Gaussian Mixture background subtraction method. \cite{Zivkovic:2004improved} is chosen for its public availability and popularity. We test it on the original $5$-second video clips. The second baseline model is a patch-based classification model using Histogram of Oriented Gradients (HOG) features \cite{Dalal:2005hog} and linear SVM. The training patches ($72 \times 72$ pixels in size) are extracted from the labeled frames in the training set. HOG features are extracted from the patches as $X$, and patch labels are the patch centers on the ground-truth segmentation maps as $y\in\{-1,1\}$. Then a linear SVM is trained with $X$ and $y$. At the testing stage, we evenly sample patches (every $10$ pixels) from the testing frames to test the trained SVM model.

In addition, we also evaluate the performances of each component of our proposed network. The three FCNN models use original frames, background subtraction frames (original frames minus the mean frame of the clip) and edge detection results from \cite{Dollar:2013structured} respectively. For data augmentation, the inputs are $256 \times 256$ pixels in size randomly cropped from original frames, background subtractions and edge model results, with $0.5$ probability of horizontal flipping. $80$\% of the scenes (161 scenes) in the training set are randomly chosen to be training scenes, and the rest (23 scenes) are using for validation, in other words, the validation set does not contain the same scenes as the training set.
% subsection baselines (end)

\subsection{System Settings} % (fold)
\label{sub:system_settings}
\vspace{0.1cm}\noindent\textbf{Data preparation and augmentation.} In total we have $6153$ frames with region-level segmentation annotation for training and validation. For data augmentation, we randomly crop $10$ $256 \times 256$ regions from each frame with $50\%$ probability of horizontal flipping. The corresponding regions on the ground-truth segmentation maps are cropped and flipped accordingly. The same augmentation process are carried out on background-subtraction samples and edge model samples.

\vspace{0.1cm}\noindent\textbf{Network Parameters.} We use a simple annotation to indicate the layer parameters: (1) \textit{Conv}(N,K,S) indicates convolutional layer with $N$ outputs, kernel size $K$ and stride size $S$. (2) \textit{Pool}(T,K,S) denotes pooling layer with type $T$, kernel size $K$ and stride size $S$. (3) \textit{ReLU} and \textit{Sig} represent rectified linear unit and sigmoid function. The our proposed FCNN can be represented as: \textit{Conv}(32,7,1) - ReLU - \textit{Pool}(MAX,2,2) - \textit{Conv}(64,7,1) - ReLU - \textit{Pool}(MAX,2,2) - \textit{Conv}(128,3,1) - ReLU - \textit{Conv}(128,3,1) - ReLU - \textit{Conv}(64,3,1) - ReLU - \textit{Conv}(16,3,1) - ReLU - \textit{Conv}(1,1,1) - Sig. All convolution operations are properly padded to keep the same shape. The segmentation ground-truth maps are pooled twice with: \textit{Pool}(AVE,2,2) - \textit{Pool}(AVE,2,2) to have the same shapes as output predictions. The loss function is cross entropy (Equation~\eqref{eq:cross_entropy}).

The three networks for appearance, motion and structure use the same network structure.

\vspace{0.1cm}\noindent\textbf{Training Strategies.} Unlike standard CNN models for classifications, the number of outputs in FCNN is usually very large ($4096$ for $64 \times 64$ output maps). The loss function is very sensitive to the initialization of parameters. We adopt a common layer-wise pre-training strategy by first training two convolutional-pooling layers and one last fusion layer. Then we add one convolutional layer at a time before last fusion layer. Finally, all the parameters are globally finetuned.

% subsection system_settings (end)

\section{Results} % (fold)
\begin{figure}[t]
\begin{center}
\includegraphics[width=\linewidth]{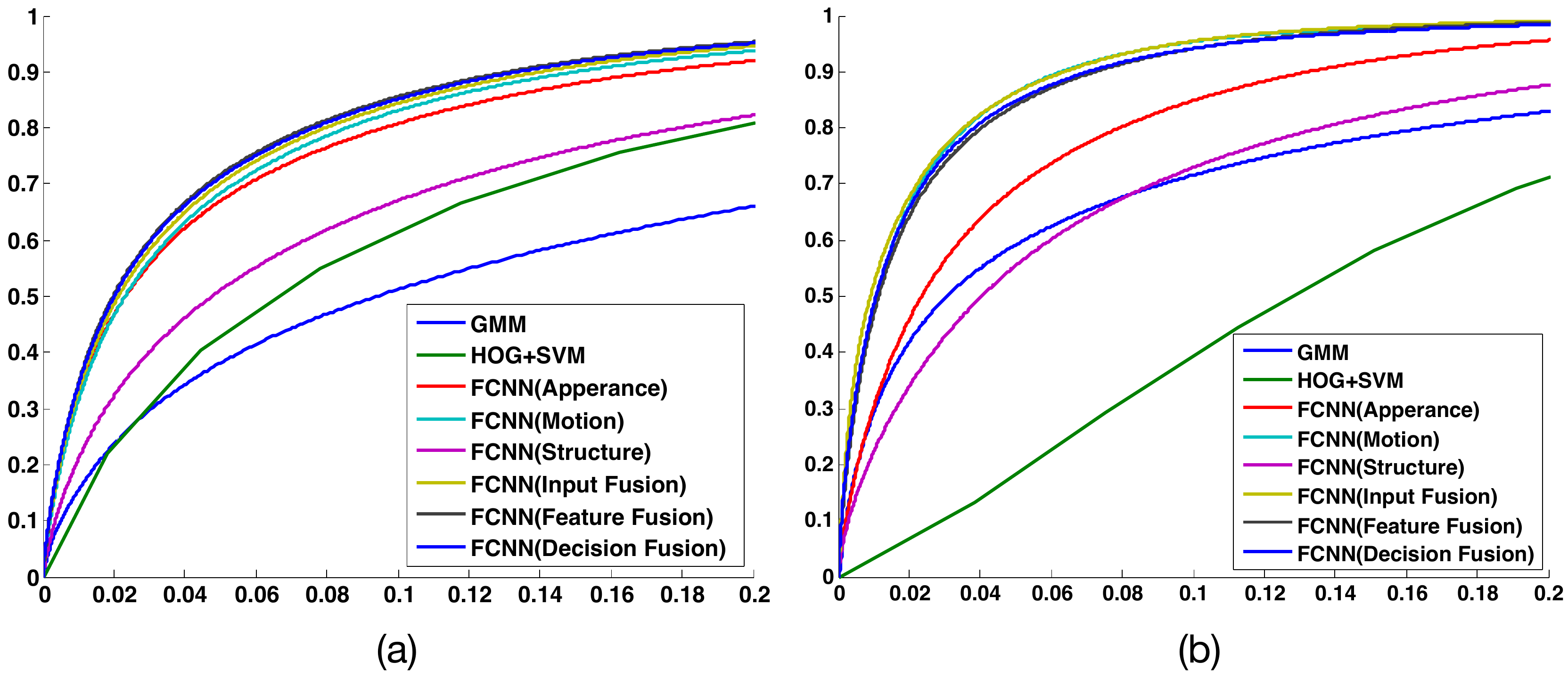}
\end{center}
   \caption{ROC curves for segmentation results on tow datasets. (a) the Shanghai World Expo dataset. (b) the City dataset.}
\label{fig:rocs}
\end{figure}

\label{sec:results}
We evaluate our method on two custom datasets that have $580$ frames with pixel-level segmentation ground truth. The two test datasets have $47$ and $11$ different scenes respectively. Figure~\ref{fig:rocs} (a) and Table~\ref{table:auc_expo} shows the ROC curves and area-under-curve (AUC) values for different methods on Shanghai World Expo dataset. Figure~\ref{fig:rocs} (b) and Table~\ref{table:auc_city} are the results on the City Dataset.

\subsection{Shanghai World Expo Dataset} % (fold)
\label{sub:shanghai_world_expo_dataset}

Figure~\ref{fig:rocs} (a) and Table~\ref{table:auc_expo} are performance on Shanghai World Expo datast. We can see that our method are better than other methods with large margin. 

\vspace{0.1cm} \noindent\textbf{Baseline models.} GMM method perform poorly on this dataset for the large number of stationary groups such as queues and squares in this dataset. In addition, the videos are $5$ second clips which are not sufficient for estimating a good background image. HOG+SVM method has reasonable performance to capture the appearance of crowded scenes. The reason is that, although the training set and testing set have no overlap in scenes, the appearance features are similar, especially in local patches. 

\vspace{0.1cm} \noindent\textbf{Single FCNN models.} The appearance model and motion model have comparable performance, with motion model slightly better. The reason is that the motion input is background subtraction without thresholding, which not only include background information, but also contains appearance information. Structure model uses edge detection results as inputs, which are similar to HOG features, and  has slightly better performance than HOG method. 

\vspace{0.1cm} \noindent \textbf{Model combinations.} The three combination schemes have comparable performances which all improves single model ability. The Input Fusion model is a single-stage combination and has slight improvement from single models. For multi-stage combinations, the main branch is appearance model, adding motion and structure information improves performance. The Feature Fusion and Decision Fusion have similar results, with Feature Fusion slightly better.

\begin{table}
\begin{center}
\small
\begin{tabular}{|l|c|}
\hline
Method & AUC \\
\hline\hline
GMM \cite{Zivkovic:2004improved} & 0.8068 \\
HOG+SVM & 0.8818 \\
\hline\hline
FCNN (Apperance) & 0.9376\\
FCNN (Motion) & 0.9430\\
FCNN (Structure) & 0.8881\\
\hline\hline
FCNN (Input Fusion) & 0.9480\\
FCNN (Feature Fusion) & \textbf{0.9511}\\
FCNN (Decision Fusion) & 0.9505\\
\hline
\end{tabular}
\end{center}
\caption{Comparison results on Shanghai World Expo Dataset.}
\label{table:auc_expo}
\end{table}

\begin{table}
\begin{center}
\small
\begin{tabular}{|l|c|}
\hline
Method & AUC \\
\hline\hline
GMM \cite{Zivkovic:2004improved} & 0.8923 \\
HOG+SVM & 0.8426 \\
\hline\hline
FCNN (Apperance) & 0.9499\\
% FCNN(Apperance+Perspective) & \\
FCNN (Motion) & 0.9739\\
FCNN (Structure) & 0.9142\\
\hline\hline
FCNN (Input Fusion) & \textbf{0.9761}\\
FCNN (Feature Fusion) & 0.9724\\
FCNN (Decision Fusion) & 0.9726\\
\hline
\end{tabular}
\end{center}
\caption{Comparison results on City Dataset.}
\label{table:auc_city}
\end{table}

\subsection{City Dataset}
Figure~\ref{fig:rocs} (b) and Table~\ref{table:auc_city} are the testing results on City Dataset. The models are trained on Shanghai World Expo training set, except GMM model which adaptively learns from target scenes.

\vspace{0.1cm}\noindent\textbf{Baseline models.} The GMM method have much better performance than that in Shanghai World Expo Dataset. This is because: 1) most people in the City Dataset are walking, therefore, it is easy to get a good background model; 2) the video clips are $1$ minute in length, which provide more temporal information for background modeling. HOG method, however, has significant performance decrease. The appearances on the two datasets are largely different and HOG features do not have a very good generalization capability. 

\vspace{0.1cm}\noindent\textbf{Single FCNN models.} In Table~\ref{table:auc_city}, all the three single FCNN models have the performance improved compared to the baselines, which shows the generalization capability of FCNN. Similar to that on Shanghai World Expo Dataset, motion model has better performance than appearance model. It is because the videos in this dataset have more frames and fewer stationary crowds. Structure model has better results than HOG models with large margin.

\vspace{0.1cm}\noindent\textbf{Model combinations.} Interestingly, the Feature Fusion and Decision Fusion models have lower, though comparable, performances than single motion model. This is because we use appearance branch as main branch in the multi-stage combination, and motion branch tends to learn complementary information to that of appearance branch. However, in single-stage combination, the motion input may be the dominating force and Input Fusion model yields the best performance.

\begin{figure*}[t]
    \centering
    \includegraphics[width=0.93\linewidth]{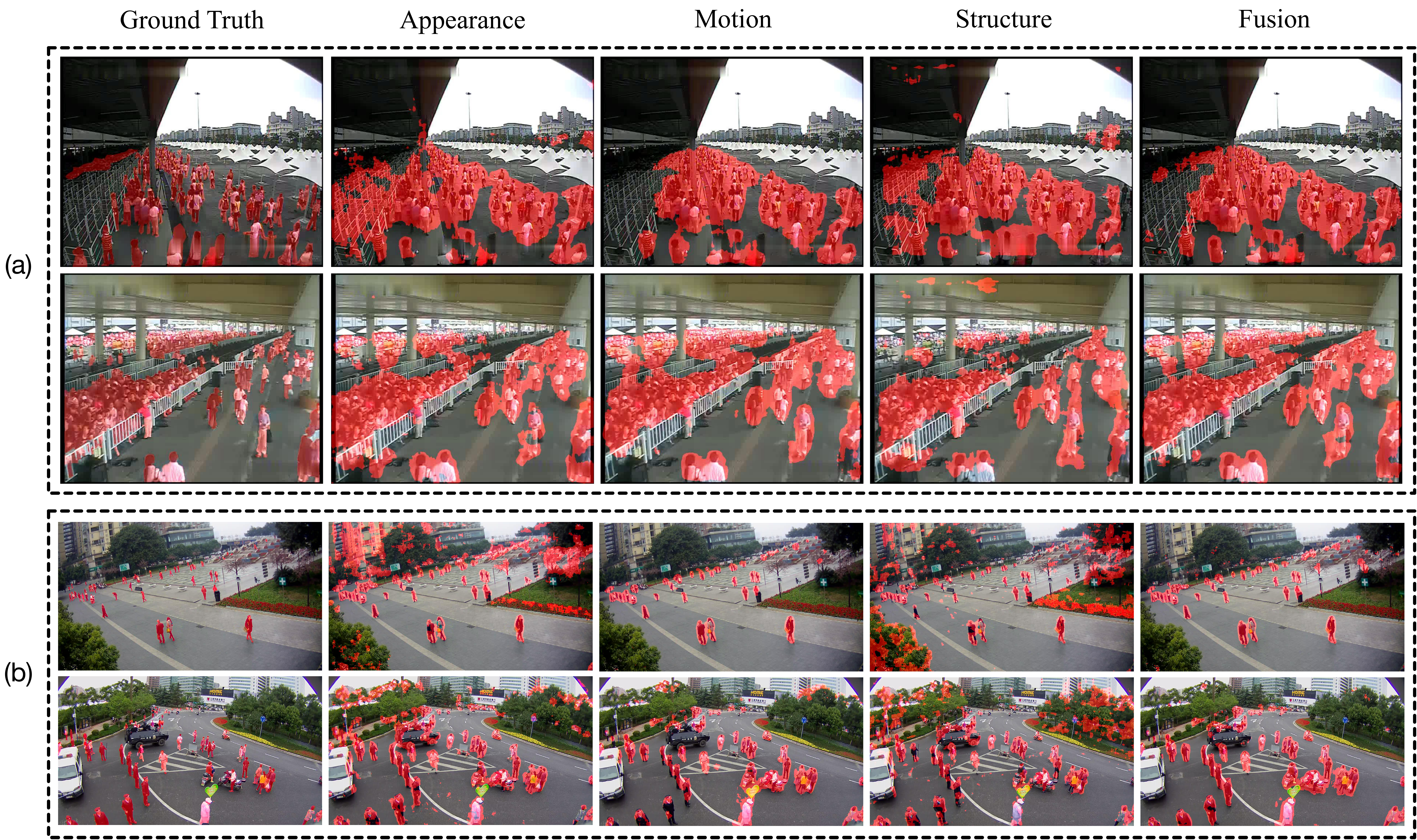}
    \caption{Fusion improvement. (a) Shanghai World Expo dataset. (b) City dataset. The appearance results contain false positives on background buildings and trees. The motion results capture moving crowds but have false negatives on stationary pedestrians. The structure results have tight segmentation contours. The fusion process combines three models to improve the segmentation results.}
    \label{fig:fusion_improvement}
\end{figure*}

\subsection{Discussions} % (fold)
\label{sub:discussions}
In Figure~\ref{fig:fusion_improvement}, we compare the results of single FCNN models with fusion models. The fusion improvements come from several sources: 1) the appearance model usually has false positives on background structures such as buildings, trees, fencing, \etc. By incorporating motion information, these false positives are removed (the first rows in Figure~\ref{fig:fusion_improvement} (a) and (b)). 2) the appearance model has false negatives in some areas far way from the camera. The motion models can capture their movements and add these areas to the segmentation results. (the second row in Figure~\ref{fig:fusion_improvement} (a)) 3) Some stationary pedestrians (the second row in Figure~\ref{fig:fusion_improvement} (b)) are missed by the motion model but are captured by the fusion model because of the appearance branch.

% subsection discussions (end)
% subsection results (end)

% section experiments (end)

\section{Conclusion} % (fold)
\label{sec:conclusion}
In this work, we propose a novel fully-convolutional neural network model for full-frame training and testing in crowd segmentation. We incorporate appearance, motion and structure information and propose three fusion schemes. To train and evaluate our model, we create two datasets that have $235$ and $11$ distinct camera views respectively. They are the largest crowd segmentation datasets up to date and will be released to the public.

% section conclusion (end)

{\small
\bibliographystyle{ieee}
\bibliography{refs}
}

\end{document}